\documentclass[aps,preprint,nofootinbib,floatfix]{revtex4-1}

\usepackage[pdftex]{graphicx}
\graphicspath{{./figures/}}
\usepackage{amsmath,amssymb,amsfonts,amsthm,amscd,bm}
\usepackage{algorithm}
\usepackage{algorithmic}

\newtheorem{theorem}{Theorem}

\newtheorem{lemma}[theorem]{Lemma}

\DeclareMathOperator{\diag}{diag}

\newcommand{\G}{\mathcal{G}}
\newcommand{\F}{\mathcal{F}}
\newcommand{\E}{\mathcal{E}}
\newcommand{\V}{\mathcal{V}}
\newcommand{\bG}{\bar{\mathcal{G}}}
\newcommand{\bF}{\bar{\mathcal{F}}}
\newcommand{\bE}{\bar{\mathcal{E}}}
\newcommand{\bV}{\bar{\mathcal{V}}}
\newcommand{\hV}{\hat{\mathcal{V}}}

\newcommand{\Dd}{D_{\bm{d}}}
\newcommand{\M}{M}
\newcommand{\hM}{\hat{M}}
\newcommand{\MC}{\mathcal{M}}
\newcommand{\hMC}{\hat{\mathcal{M}}}
\newcommand{\HH}{\mathcal{H}}
\newcommand{\hHH}{\hat{\mathcal{H}}}
\newcommand{\vpi}{\bm{\pi}}
\newcommand{\hvpi}{\hat{\bm{\pi}}}

\newcommand{\A}{A}
\newcommand{\B}{B}

\newcommand{\vx}{\bm{x}}
\newcommand{\vz}{\bm{z}}
\newcommand{\vu}{\bm{u}}
\newcommand{\vm}{\bm{m}}
\newcommand{\vn}{\bm{n}}
\newcommand{\vs}{\bm{s}}
\newcommand{\vv}{\bm{v}}

\newcommand{\vone}{\bm{1}}
\newcommand{\vzero}{\bm{0}}
\newcommand{\vrho}{\bm{\rho}}

\newcommand{\R}{\mathcal{R}}

\newcommand{\HA}{\mathcal{R}_A}
\newcommand{\HG}{\mathcal{R}_G}
\newcommand{\hx}{\hat{x}}
\newcommand{\hy}{\hat{y}}
\newcommand{\hz}{\hat{z}}

\begin{document}

\title{Markov Chain Lifting and Distributed ADMM}

\author{Guilherme Fran\c ca}
\email{guifranca@gmail.com}
\affiliation{Boston College, Computer Science Department}
\affiliation{Johns Hopkins University, Center for Imaging Science}

\author{Jos\' e Bento}
\email{jose.bento@bc.edu}
\affiliation{Boston College, Computer Science Department}

\begin{abstract}
The time to converge to the steady state of a finite Markov chain can be
greatly reduced by a lifting operation,
which creates a new Markov chain on an expanded state space. For a class of
quadratic objectives, we show an analogous behavior where a distributed 
ADMM algorithm can be seen as a lifting of
Gradient Descent algorithm. 
This provides a deep insight for its faster convergence
rate
under optimal parameter tuning.
We conjecture that this gain is always present,
as opposed to the lifting of a Markov chain which sometimes
only provides a marginal speedup.
\end{abstract}

\maketitle

\section{Introduction}
\label{sec:intro}

\begin{figure}[t]
\centering
\includegraphics[scale=1.2]{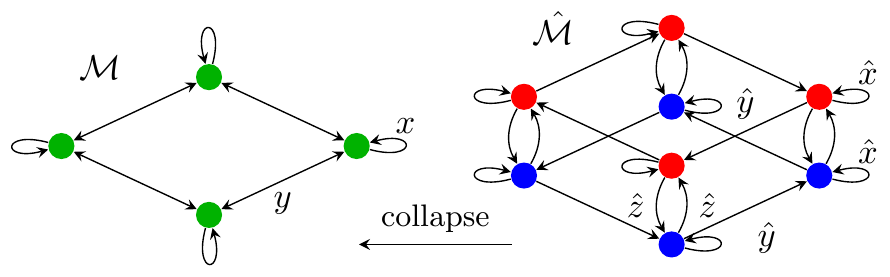}
\caption{The base Markov chain $\MC$ over the cycle graph $C_4$  and
its lifted Markov chain $\hMC$ with duplicate states. Each 
state in $\V$ (green nodes) is expanded into 
two states in $\hV$ (blue and red nodes above each other).
}
\label{fig:ring_lifting}
\end{figure}

Let $\MC$ and $\hMC$ be two finite Markov chains with 
states $\V$ and $\hV$, of sizes $|\V| < |\hV|$, and with
transition matrices $\M$ and $\hM$, respectively.
Let their stationary distributions be $\vpi$ and $\hvpi$.
In some cases it is possible to use $\hMC$ to sample 
from the stationary distribution of $\MC$.
A formal set of conditions under which this happen is 
known as \emph{lifting}.
We say that $\hMC$ is a lifting of $\MC$ 
if there is a row stochastic matrix
$S\in \mathbb{R}^{|\hat{\V}| \times |\V|}$ with 
elements $S_{ij} \in \{0,1\}$ and a single nonvanishing element
per line, where $\vone_{|\V|} = S^\top \vone_{|\hV|}$,  and
$\vone_n$ is the all-ones 
$n$-dimensional vector, such that
\begin{equation}\label{eq:lifting_def}
\bm{\pi} = S^\top \hat{\bm{\pi}}, \qquad
D_{\bm{\pi}} M = S^\top D_{\hat{\bm{\pi}}} \hat{M} S.
\end{equation}
We denote ${S}^{\top}$ the transpose of $S$, and
for any vector $\vv \in \mathbb{R}^n$, $D_{\vv} = \diag(v_1,\dots,v_n)$.
Intuitively, $\hMC$ contains copies of the states of $\MC$ 
and transition probabilities between the extended sates $\hV$ such that
it is possible
to collapse $\hMC$ onto $\MC$. This is the meaning of relation
\eqref{eq:lifting_def}. See 
Fig.~\ref{fig:ring_lifting} for an illustration.
(We refer to \cite{chen1999lifting} 
for more details on Markov
chain lifting.)

The \emph{mixing time} $\HH$ is a measure of the time it takes
for the distribution of a Markov chain $\MC$ to approach stationarity.
We follow the definitions of \cite{chen1999lifting} 
but, up to multiplicative factors and slightly loser bounds, 
the reader can think of
\begin{equation}
\label{eq:mixing}
\HH = \min \{t:  \max_{\{i, \bm{p}^0\}} | p^{t}_i  - \pi_i | < 1/4 \},
\end{equation}
where $p^{t}_i$ is the probability of being on state $i$ 
after $t$ steps, starting from the initial distribution $\bm{p}^0$.
Lifting is particularly useful when the mixing time
$\hHH$ of
the lifted chain is much smaller than $\HH$.
There are several examples where $\hHH \approx C \sqrt{\HH}$, for
some constant $C \in (0,1)$ which depends only on $\bm{\pi}$.
However,  there is a limit on how much speedup can be achieved.
If $\MC$ is irreducible, then
$\hHH \ge C \sqrt{\HH}$. 
If $\MC$ and $\hMC$ are reversible, then
the limitation is even stronger,
$ \hHH \ge C \HH$. 

Consider the  
undirected and connected  graph
$\G = (\V, \E)$, with vertex set $\V$ and edge set $\E$.
Let $\vz \in \mathbb{R}^{|\V|}$ with components 
$z_i$, and consider the quadratic problem
\begin{equation}
\label{eq:general_quad_obj}
\min_{ \vz \in \mathbb{R}^{|\V|}} \Big \{ f(\bm{z}) = 
\dfrac{1}{2} \sum_{(i,j)  \in \E} 
q_{ij} (z_i - z_j)^2 \Big \} .
\end{equation}
We also write $q_{ij} = q_e$ for $e = (i,j) \in \E$. 
There is a connection between solving \eqref{eq:general_quad_obj} by
\emph{Gradient Descent} (GD) algorithm and the evolution of a Markov chain.
To see this, consider $q_e = 1$ for simplicity.
The GD iteration with step-size $\alpha > 0$ is given by
\begin{equation}\label{eq:GD_rule}
\vz^{t+1} = \left( I - \alpha \nabla f  \right) \vz^{t} 
= \left( I - \alpha \Dd (I - W) \right) \vz^{t}
\end{equation}
where $W = \Dd^{-1} A$ is the
transition matrix of a random walk on $\G$,
$A$ is the adjacency matrix, and $\Dd$ is the degree
matrix, where
$\bm{d} = \diag(d_1,\dotsc,d_{|\V|})$.

This connection is specially clear for $d$-regular graphs.
Choosing $\alpha = 1/d$, equation \eqref{eq:GD_rule} simplifies to
$\vz^{t+1} =  W \vz^{t}$.
In particular, the \emph{convergence rate} of GD is determined by the
spectrum of
$W$, which is connected to the mixing time of $\MC$. More precisely,
when $W$ is irreducible and aperiodic, and denoting
$\lambda_2(W)$ its second largest eigenvalue in absolute value, the
mixing time of the Markov chain 
and the \emph{convergence time} of GD 
are both \mbox{equal to}
\begin{equation}
\label{eq:convergence_time_and_eigenvalues}
\mathcal{H} = \dfrac{C}{\log (1/|\lambda_2|)} 
\approx \dfrac{C}{1 -  |\lambda_2|}
\end{equation}
where
the constant $C$ comes from the tolerance error, which in 
\eqref{eq:mixing} is $1/4$. In the above approximation we assumed
$\lambda_2 \approx 1$.
Therefore, at least for GD, we can use the theory of
Markov chains to analyze the convergence rate when solving 
optimization problems.
For 
this example, 
and whenever there is linear convergence, the convergence rate $\tau$ 
and the convergence
time $\mathcal{H}$ are related by $\tau^\mathcal{H} = \Theta(1)$.
For an introduction on Markov chains, mixing times, and transition
matrix eigenvalues, we refer the reader to \cite{Norris}.

\begin{figure}
\centering
\includegraphics[scale=1.0]{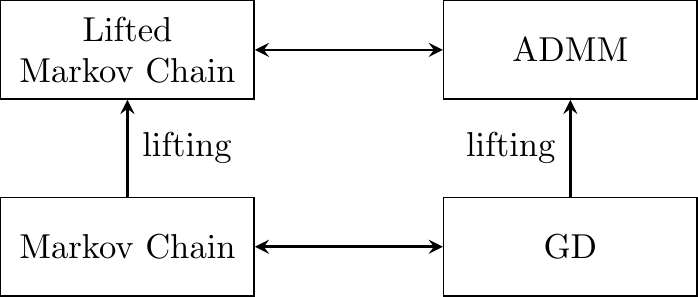}
\caption{
GD is the analogue of a Markov chain, while
distributed ADMM is the analogue of a lifted version of 
this Markov chain, which mixes faster.
}
\label{fig:scheme}
\end{figure}

The main goal of this paper is to extend the above connection
to the \emph{over-relaxed Alternating Direction Method of Multipliers} (ADMM)
algorithm, 
and the concept of lifting will play an important role.
Specifically, for problem \eqref{eq:general_quad_obj}, 
we show that a distributed implementation of over-relaxed 
ADMM can be seen as a lifting
of GD, in the same way that $\hMC$ is a lifting of the Markov chain $\MC$.
More precisely, there is a matrix $M_A$ with stationary vector
$v_A$ associated to distributed ADMM, and a matrix $M_G$ with stationary
vector $v_G$ associated to GD, such that 
relation \eqref{eq:lifting_def} is satisfied.
This duality is summarized in Fig.~\ref{fig:scheme}.
In some cases, $M_A$ might have a few negative entries
preventing it from being the transition matrix of a Markov chain.
However, it always satisfies all the other properties of
Markov matrices.

As explained in the example preceding
\eqref{eq:convergence_time_and_eigenvalues},
the convergence time of an algorithm can be related to 
the mixing time of a Markov chain. Let
$\HH_A$ be the convergence time of ADMM, and $\HH_G$ the convergence
time of GD. The lifting  relation between both algorithms strongly
suggest that, 
for problem
\eqref{eq:general_quad_obj} and optimally tuned parameters, 
ADMM is always faster than GD.
Since lifting can speed
mixing times of Markov chains up to a square root factor, 
we conjecture that
the optimal convergence  times
$\HH^\star_A$ and 
$\HH^\star_G$ are related as
\begin{equation}
\label{eq:conj}
\HH^*_A \le C \sqrt{\HH^*_G},
\end{equation}
where $C > 0$ is a universal constant. 
Moreover, we conjecture that \eqref{eq:conj} holds for \emph{any}
connected graph $\G$.
Note that \eqref{eq:conj} is much stronger 
than the analogous relation for 
lifted
of Markov chains, 
where for some graphs, e.g. with low conductance, the gain
is marginal.

The outline of this paper is the following.
After mentioning related works in Section~\ref{sec:related_work}, 
we state our main results in Section~\ref{sec:ADMMLifting},
which shows that distributed implementations of
over-relaxed ADMM and GD obey the lifting
relation  \eqref{eq:lifting_def}.
The proofs can be found in the 
Appendix. In Section~\ref{sec:numerical}
we support conjecture~\eqref{eq:conj} with numerical evidence.
We present our final remarks in Section~\ref{sec:conclusion}.

\section{Related Work and an Open Problem}
\label{sec:related_work}

We state conjecture~\eqref{eq:conj} 
for the relatively simple problem~\eqref{eq:general_quad_obj}, 
but, to the best of our knowledge, 
it cannot be resolved through the existing literature.
We compare the \emph{exact
asymptotic convergence rates after optimal tuning}
of ADMM and GD, while
the majority of previous papers focus on upper bounding
the \emph{global convergence rate} of ADMM
and, at best, optimize such an upper bound.

Furthermore, 
to obtain linear convergence, strong convexity is usually 
assumed \cite{shi2014linear},
which does not hold for problem 
\eqref{eq:general_quad_obj}. 
Most results not requiring strong convexity focus on the convergence 
rate of the objective function, as opposed to this paper
which focus on the convergence rate of 
the variables; see 
\cite{davis2014convergence}
for example.

Few papers consider a consensus problem with
an objective function different than \eqref{eq:general_quad_obj}.
For instance, 
\cite{erseghe2011fast} considers 
$f(\vz) = \sum_{i \in \V }\sum\|z_i - c_i\|^2$,
subject to  $z_i = z_j$ if $(i,j) \in \E$, where $c_i > 0$ are 
constants. This problem is strongly convex and does not reduce to
\eqref{eq:general_quad_obj}, and vice-versa.
Other branch of research consider $f(\vz) = \sum_i f_i(\vz)$ with
ADMM iterations that are agnostic to 
whether or not $f_i(\vz)$ depends on a subset 
of the components of $\vz$; see 
\cite{wei2012distributed} and references therein.
These are in contrast with our setting where
decentralized ADMM is a message-passing algorithm \cite{Bento1},
and the messages between agents $i$ and $j$ 
are only associated to the variables shared by functions
$f_i$ and $f_j$.

For quadratic problems, 
there are explicit results on the convergence rate and
optimal parameters of ADMM
\cite{teixeira2013optimal,ghadimi2015optimal,iutzeler2016explicit}.
However, their assumptions do not hold 
for the non strongly convex distributed problem considered in this paper.
Moreover, there are very few results comparing
the optimal convergence rate of ADMM as a function of
the optimal convergence rate of GD. For a centralized
setting, an explicit comparison is
provided in \cite{FrancaBento}, but it assumes
strong convexity.

Finally, and most importantly, there is no prior result 
connecting 
GD and ADMM to lifted Markov chains.
Lifted Markov chains were previously employed 
to speedup convergence time
of 
distributed 
averaging and gossip algorithms 
\cite{jung2007fast, li2010location, jung2010distributed}, but these
do not involve ADMM algorithm.

\section{ADMM as a Lifting of Gradient Descent}
\label{sec:ADMMLifting}

We now show that the lifting relation \eqref{eq:lifting_def} 
holds when distributed implementations of over-relaxed
ADMM and GD are applied to 
problem \eqref{eq:general_quad_obj} defined over the graph $\G = (\V, \E)$.

Let us introduce the extended set 
of variables $\vx \in \mathbb{R}^{|\hat{\E|}}$,
where 
\begin{equation}
\hat{\E} = \{ (e,i): e\in \E, \, i \in e, \, \text{and} \, i \in \V \}.
\end{equation}
Note that
$|\hat{\E}| = 2|\E|$. Each component of $\vx$ is indexed
by a pair $(e,i) \in \hat{\E}$. For simplicity we denote $e_i=(e,i)$. 
We can now write \eqref{eq:general_quad_obj} as
\begin{equation}
\label{eq:fTilde}
 \min_{\vx,\vz}  
 \Big \{ f({\vx}) = 
\dfrac{1}{2} \hspace{-.2cm}
\sum_{e=(i,j) \in \E} \hspace{-.2cm}
q_{e} (x_{e_i} - x_{e_j})^2 \Big \} 
\end{equation}
%
subject to $x_{e_i} = z_i$, $x_{e_j} = z_j$,    
for all $e = (i,j) \in \E$.
The new variables are defined according to the following
diagram:
\begin{figure}[h]
\centering
\includegraphics[scale=1.2]{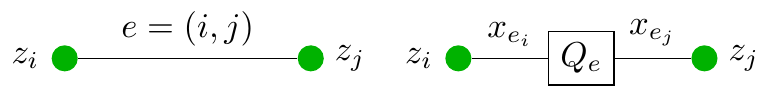}
\end{figure}\\
Notice that we can also write ${f}(\vx) = \tfrac{1}{2} {\vx}^\top Q \vx$
where $Q$ is block diagonal, one block per edge $e = (i,j)$, in the 
form $Q_e = q_{e} \left( \begin{smallmatrix} +1  & -1 \\ -1 & +1
\end{smallmatrix}\right) $.
Let us define the matrix $S \in \mathbb{R}^{|\hat{\E}| \times|\V|}$ with
components 
\begin{equation}
S_{e_i,i} = S_{e_j,j} = \begin{cases}
1 & \mbox{if $e = (i,j)\in \E$,} \\
0 & \mbox{otherwise.}
\end{cases}
\end{equation}

The distributed over-relaxed ADMM is a first order method
that operates on five variables: $\vx$ and $\vz$ 
defined above, and also $\vu$, $\vm$ and $\vn$ introduced
below. It depends on  
the relaxation parameter $\gamma \in (0,2)$, 
and several penalty parameters $\vrho \in \mathbb{R}^{|\hat{\E}|}$. 
The components of $\vrho$ are $\rho_{e_i} > 0$ 
for $e_i \in \hat{\E}$; see \cite{FrancaBento, Bento1} and
also \cite{Bento2}
for details on multiple $\rho$'s.
We can now write ADMM iterations as
\begin{equation} \label{eq:admm_matrix}
\begin{aligned}
\bm{x}^{t+1} &=  \A \bm{n}^t,  \\
\bm{m}^{t+1} &= \gamma \bm{x}^{t+1} + \bm{u}^t, \\
\bm{s}^{t+1} &= (1-\gamma) \bm{s}^t + \B \bm{m}^{t+1}, \\
\bm{u}^{t+1} &= \bm{u}^t + \gamma \bm{x}^{t+1} + (1-\gamma) \bm{s}^t - 
\bm{s}^{t+1}, \\
\bm{n}^{t+1} &= \bm{s}^{t+1} - \bm{u}^{t+1},
\end{aligned}
\end{equation}
where $\bm{s}^t = S \bm{z}^t$, 
$\B = S (S^\top D_{\bm{\rho}} S)^{-1} S^\top D_{\bm{\rho}} $,
and $\A = (I + D^{-1}_{\bm{\rho}} Q)^{-1}$.  
The next result shows that these iterations are equivalent
to a linear system in $|\hat{\E}|$ dimensions. (The proofs of the following
results are in the Appendix.)
\begin{theorem}[Linear Evolution of ADMM] \label{th:linearADMM}
Iterations \eqref{eq:admm_matrix}
are equivalent to
\begin{equation}\label{eq:TA}
\bm{n}^{t+1} = T_A \, \bm{n}^t,  \qquad  
T_A = I - \gamma(\A + \B - 2 \B\A),
\end{equation}
with $\bm{s}^t = B \bm{n}^t$ and $\bm{u}^t = - (I - B) \bm{n}^t$.
All the variables of ADMM depend only on $\vn^t$.
\end{theorem}

We can also write GD update $\vz^{t+1} = (I - \alpha \nabla f)\vz^t$ 
to problem \eqref{eq:general_quad_obj} as
\begin{equation}
\label{eq:TG2}
\vz^{t+1} = T_G \vz^t,
\qquad T_G = I - \alpha S^\top Q S.
\end{equation}

In the following, we establish lifting relations
between distributed ADMM and GD in terms of matrices $M_A$ and $M_G$, which 
are very closely related but not necessarily equal to $T_A$ 
and $T_G$. They are defined as
\begin{align}
M_G &= (I - D_G)^{-1}(T_G - D_G), \label{eq:MAMG1} \\
M_A &= (I - D_A)^{-1}(T_A - D_A), \label{eq:MAMG2}
\end{align}
where $D_G \ne I$ and $D_A\ne I$ are, for the moment, 
arbitrary diagonal matrices.
Let us also introduce the vectors
\begin{align}
\bm{v}_G &= (I - D_G) \bm{1}, 
\label{eq:vg}\\
\bm{v}_A &= (I - D_A) \bm{\rho}.
\label{eq:va}
\end{align}

As shown below, these matrices and vectors satisfy
relation \eqref{eq:lifting_def}. Moreover,
$M_G$ can be interpreted as a probability transition matrix,
and the rows of $M_A$ sum up to one.
We only lack the strict non-negativity of $M_A$, which in general 
is not a probability transition matrix. Thus, in general, 
we do not have a lifting between Markov chains, however, we still have
lifting in the sense that $M_A$ can be collapsed onto $M_G$ according
to \eqref{eq:lifting_def}.

\begin{theorem}\label{th:TG_stochastic}
For $(D_G)_{ii} < 1$ and  sufficiently small $\alpha$, 
$M_G$ in \eqref{eq:MAMG1} is a 
doubly stochastic matrix. 
\end{theorem}

\begin{lemma}\label{th:TAoneisone}
The rows of $M_G$ and $M_A$ sum up to one, i.e. $M_G\vone = \vone$ and
$M_A \vone = \vone$. Moreover,
${\vv}_G^\top M_G = {\vv}_G^\top$ 
and $\vv_A^\top M_A = \vv_A^\top$. These properties are shared with
Markov matrices.
\end{lemma}

\begin{theorem}[ADMM as a Lifting of GD]
\label{th:lifting}
$M_A$ and $M_G$ defined in \eqref{eq:MAMG1} and \eqref{eq:MAMG2} 
satisfy relation \eqref{eq:lifting_def}, namely,
\begin{equation} 
\label{eq:liftingMAMG}
\bm{v}_G = S^\top \bm{v}_A, \qquad
D_{\bm{v}_G} M_G = S^\top D_{\bm{v}_A} M_A S ,
\end{equation}
provided $D_G$, $D_A$, $\alpha$, $\gamma$, and $\vrho$ are related
according to
\begin{align}
&S^\top D_{\bm \rho} (I - D_A) S = I - D_G ,
\label{eq:lifting_params} \\
&\alpha = \dfrac{\gamma \, q_e \rho_{e_i, i} \, \rho_{e_j, j} }{
\rho_{e_i, i} \, \rho_{e_j, j} + 
q_e\left(\rho_{e_i, i} + \rho_{e_j ,j} \right)}, 
\label{eq:rho_alpha}
\end{align}
for all $e = (i,j) \in \E$. 
Equation \eqref{eq:rho_alpha} restricts the components of $\vrho$, and
\eqref{eq:liftingMAMG} is an equation for 
$D_A$ and $D_G$.
\end{theorem}

\begin{theorem}[Negative Probabilities]\label{th:negative}
There exists a graph $\G$ such that, 
for any diagonal matrix $D_A$, $\vrho$ and $\gamma$, the matrix $M_A$ 
has at least one negative entry.
Thus, in general, $M_A$ is not a probability transition matrix.
\end{theorem}

For concreteness, let us consider some explicit examples illustrating 
Theorem~\ref{th:lifting}.

\paragraph*{\bf Regular Graphs.}
\label{regular}
Let us consider the solution to
equations
\eqref{eq:lifting_params} and
\eqref{eq:rho_alpha}
for  $d$-regular graphs.
Fix $q_e = 1$ and $\vrho = \rho \vone$ for simplicity.
Equation \eqref{eq:lifting_params} is satisfied with
\begin{equation}
\label{eq:regular_DA_DG}
D_A = (1 -  (\rho |\hat{\E}|)^{-1} ) I, \qquad
D_G = (1 -  |\V|^{-1} )I, 
\end{equation}
since $d |\V| = |\hat{\E}| = 2 |\E|$, while  
\eqref{eq:rho_alpha} requires
\begin{equation}
\label{eq:regular_alpha_rho}
\alpha = \dfrac{\gamma \rho}{2+ \rho}.
\end{equation}
Notice that $(D_G)_{ii} < 1$ for all $i$, 
so choosing $\gamma$ or $\rho$ small enough 
we can make $M_G$ positive. Moreover, the components of
 \eqref{eq:vg} and \eqref{eq:va} are non-negative and
sum up to one, i.e.
${\vv}^\top_G {\bm 1} = {\vv}^\top_A {\bm 1} = 1$,
thus these vectors are stationary probability distributions 
of $M_G$ and $M_A$.

\paragraph*{\bf Cycle Graph.}
\label{ring}
Consider solving 
\eqref{eq:general_quad_obj} over the
$4$-node cycle graph $\G = C_4$ shown in 
Fig.~\ref{fig:ring_lifting}. 
By direct computation and upon using
\eqref{eq:regular_DA_DG} we obtain
\begin{equation}
M_G = \begin{pmatrix}
x & y & 0 & y\\
y & x & y & 0\\
0 & y & x & y\\
y & 0 & y & x\\
\end{pmatrix}, \qquad
M_A = 
\begin{pmatrix}
\hx & 0 & 0 & 0 & 0 & 0 & \hy & \hz \\
0 & \hx & \hz & \hy & 0 & 0 & 0 & 0 \\
\hy & \hz & \hx & 0 & 0 & 0 & 0 & 0 \\
0 & 0 & 0 & \hx & \hz & \hy & 0 & 0 \\
0 & 0 & \hy & \hz & \hx & 0 & 0 & 0 \\
0 & 0 & \hy & 0 & 0 & \hx & \hz & \hy \\
0 & 0 & 0 & 0 & \hy & \hz & \hx & 0 \\
\hz & \hy & 0 & 0 & 0 & 0 & 0 & \hx
\end{pmatrix},
\end{equation}
where the probabilities of $M_G$ are given by
$x = 1 -8\alpha$ and $x + 2y = 1$, and the probabilities of $M_A$
are
$\hx = 1 - 4\gamma \rho$, $\hy = 8\gamma \rho / (2+\rho)$ and
$\hx + \hy + \hz = 1$.
The stationary probability vectors are
$\vpi = \tfrac{1}{4}\vone$ and
$\hat{\vpi} = \tfrac{1}{8}\vone$. Now 
\eqref{eq:liftingMAMG} holds provided the parameters are related
as \eqref{eq:regular_alpha_rho}.
Moreover, in this particular case the matrix $M_A$ is strictly
non-negative, thus ADMM is a lifting of GD in the Markov chain sense.

\bigskip

Based on the above theorems we propose conjecture~\eqref{eq:conj}.
The convergence rate $\tau$ is related to the convergence time, for instance
$\HH \sim (1-\tau)^{-1}$ if $\tau \approx 1$. Thus,
let $\tau_G^\star$ and $\tau_A^\star$ be the optimal convergence rates
of GD and ADMM, respectively. Then, 
at least for objective \eqref{eq:general_quad_obj},
and for any $\G$,  we conjecture that
there is some universal constant $C>0$ such that
\begin{equation}
\label{eq:conjecture_tau}
1-\tau_A^\star \ge C \sqrt{1-\tau_G^\star}.
\end{equation}
%

\section{Numerical evidence}
\label{sec:numerical}

For many graphs, we observe very few negative entries
in $M_A$, which can be further
reduced by adjusting the parameters $\bm{\rho}$ and $\gamma$.
Nonetheless, in general, the lack of strict non-negativity of $M_A$
prevents us from directly applying
the theory of lifted Markov chains to prove \eqref{eq:conjecture_tau}.
However, there is compelling numerical evidence to
\eqref{eq:conjecture_tau} as we now show.

Consider a sequence of graphs $\{\G_n\}$, where $n = |\V|$,
such that $\tau^\star_G \to 1$ and $\tau^\star_A \to 1$ as $n\to \infty$.
Denote 
$\HG(n)  = (1-\tau_G^\star)^{-1}$ and
$\HA(n)  = (1-\tau_A^\star)^{-1}$.
We look for the smallest
$\beta$ such that
$\HA(n) \leq C \, \HG(n)^\beta$, for some $C>0$, and all large enough $n$.
If \eqref{eq:conjecture_tau} was false, there would exist sequences
$\{ \G_n\}$ for which
$\beta > 1/2$. For instance, if $\{\G_n\}$
have low conductance it is well-known that lifting does not speedup 
the mixing time  
and 
we could find $\beta = 1$.

To numerically find $\beta$, we plot 
\begin{equation} \label{eq:betas}
\hat{\beta}_1 = \dfrac{\log\HA(n)}{\log\HG(n)} \quad \mbox{and} \quad
\hat{\beta}_2 = \dfrac{\HG(n)}{\HA(n)} \,
\dfrac{\Delta \R_A(n)}{\Delta \R_G(n)} 
\end{equation}
against  $n$, 
where 
$\Delta h (n) = h(n+1) - h(n)$
for any function $h(n)$. 
The idea behind this is very simple.
Let $f(x) = C g(x)^\beta$, and $f, g \to \infty$ as $x\to\infty$. Then,
${\log f} / {\log g} \to \beta$ 
and also
${\partial_x \log f} / {\partial_x \log g} = 
{( g \, \partial_x f )} / { (f \, \partial_x g )} \to \beta$ as
$x\to\infty$. 
Thus, we analyze \eqref{eq:betas} which are 
their discrete analogue. Given a graph $\G_n$, 
from \eqref{eq:TG2} and \eqref{eq:TA} we numerically compute 
$\tau = \max_j\{|\lambda_j(T)| : |\lambda_j(T)| < 1 \}$.
The optimal convergence rates are thus given by
$\tau_G^\star = \min_{\alpha} \tau_G$ and 
$\tau_A^\star = \min_{\{\gamma,\rho \}} \tau_A$, where
we consider $\vrho = \rho{\bm 1}$ with $\rho > 0$.

\begin{figure}
\begin{minipage}{.32\textwidth}
\centering
\vspace{.5cm}
\includegraphics{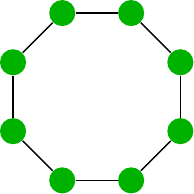}
\put(-35,-18){(a)}
\end{minipage}
\begin{minipage}{.32\textwidth}
\centering
\includegraphics{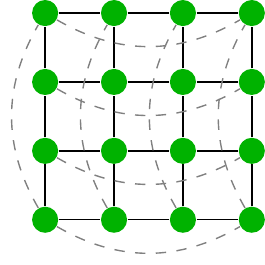}
\put(-40,-12){(b)}
\end{minipage}
\begin{minipage}{.32\textwidth}
\vspace{.5cm}
\centering
\includegraphics[angle=90]{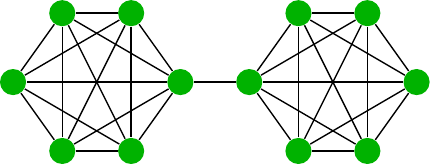}
\put(-68,-27){(c)}
\end{minipage}
\caption{
We consider the following graphs for numerical analysis. (a) Cycle graph,
$\G = C_n$. (b) Torus or periodic grid graph, $\G = T_n = C_{\sqrt{n}}\times
C_{\sqrt{n}}$. (c) Barbel graph, obtained by connecting two complete graphs
$K_n$ by a bridge.
}
\label{fig:graph_types}
\end{figure}

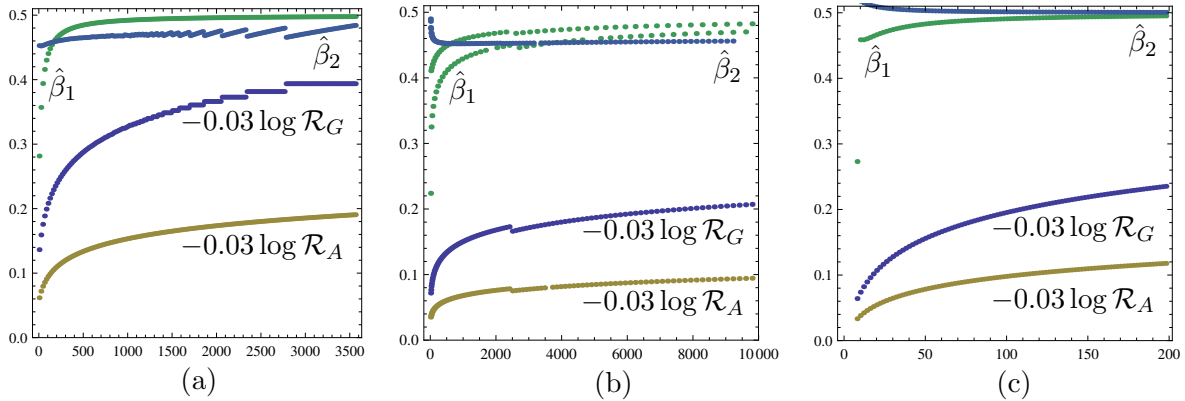
\begin{figure}
\begin{minipage}{.32\textwidth}
\centering
\includegraphics[width=.9\textwidth]{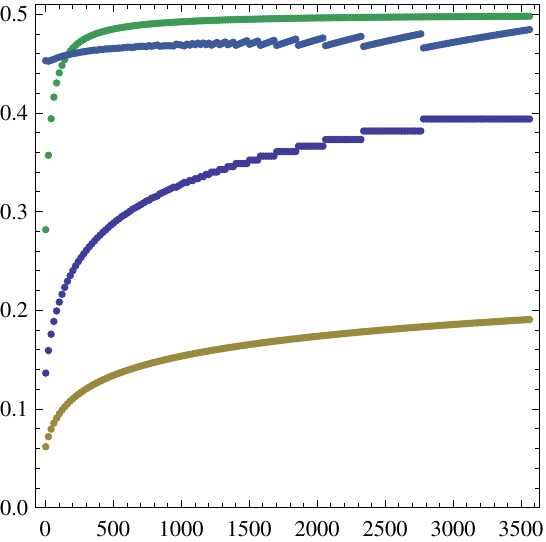}
\put(-120,100){\small $\hat{\beta}_1$}
\put(-20,112){\small $\hat{\beta}_2$}
\put(-70,40){\small $-0.03 \log \HA$}
\put(-70,87){\small $-0.03 \log\HG$}
\put(-70,-11){(a)}
\end{minipage}%
\begin{minipage}{.33\textwidth}
\centering
\includegraphics[width=.93\textwidth]{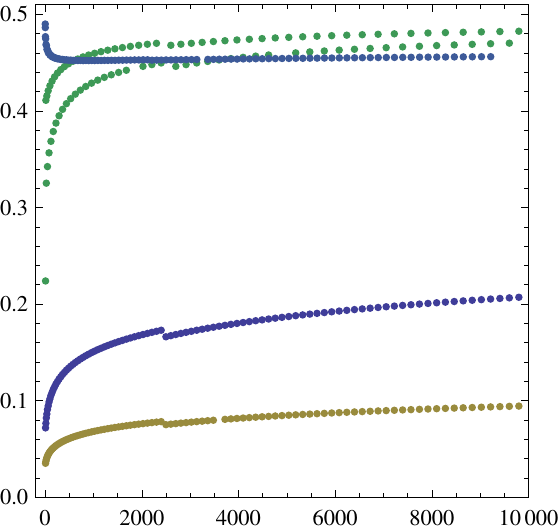}
\put(-125,100){\small $\hat{\beta}_1$}
\put(-25,108){\small $\hat{\beta}_2$}
\put(-75,20){\small $-0.03 \log \HA$}
\put(-75,47){\small $-0.03 \log\HG$}
\put(-70,-11){(b)}
\end{minipage}%
\begin{minipage}{.33\textwidth}
\centering
\includegraphics[width=.9\textwidth]{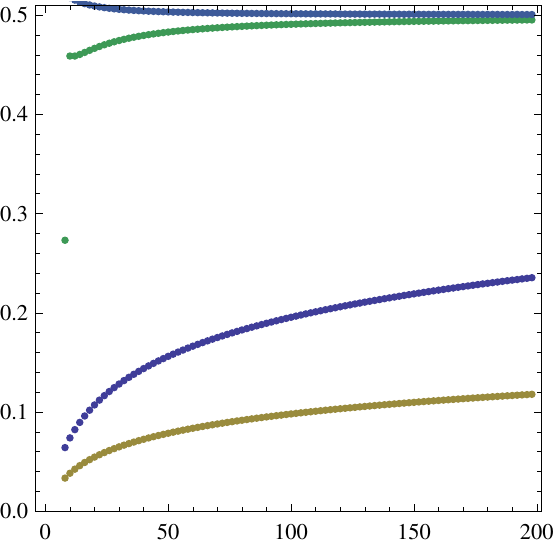}
\put(-120,110){\small $\hat{\beta}_1$}
\put(-20,118){\small $\hat{\beta}_2$}
\put(-72,21){\small $-0.03 \log \HA$}
\put(-72,48){\small $-0.03 \log\HG$}
\put(-70,-11){(c)}
\end{minipage}%
\caption{Plot of \eqref{eq:betas} versus $n$, and also $\log \HA$ and $\log \HG$
versus $n$, scaled by $-0.03$ for visualization purposes only.
The sequence of graphs $\{\G_n\}$ in each plot are of the types indicated
in Fig.~\ref{fig:graph_types}, in the same respective order.
Notice that $\hat{\beta}_1,\hat{\beta}_2 \leq 1/2$, 
and $\hat{\beta}_1$ and $\hat{\beta_2}$ gets very close to $1/2$ for large $n$.
(a) Cycle graph. (b) Torus grid graph.
The two green curves occur 
because odd and even $n$ behave differently.
(c) Barbell graph. 
A Markov chain over this graph does not speedup via lifting, however,
\eqref{eq:conjecture_tau} still holds.
}
\label{fig:betas_ring}
\end{figure}

In Fig.~\ref{fig:graph_types} we show the three different graphs considered
in the numerical analysis contained in the respective plots of 
Fig.~\ref{fig:betas_ring}. We show the values of
\eqref{eq:betas} versus $n$, and also
the curves $\log \HA$ and $\log \HG$ against $n$, 
which for visualization purposes are scaled by the factor $-0.03$.
In Fig.~\ref{fig:betas_ring}a we
see that \eqref{eq:conj}, or equivalently \eqref{eq:conjecture_tau}, 
holds for the cycle graph $\G = C_n$.
The same is true for the 
periodic grid, or torus grid graph $\G = T_{n}$, 
as shown in Fig.~\ref{fig:betas_ring}b.
Surprisingly, as shown in Fig.~\ref{fig:betas_ring}c,
we get the same square root speedup for a barbell graph,
whose random walk is known to not speedup via  lifting.
We find similar behavior for several other graphs but we omit these
results due to the lack of space.

\section{Conclusion}
\label{sec:conclusion}

For a class of quadratic problems \eqref{eq:general_quad_obj} we
established a duality between lifted Markov chains and two important
distributed optimization algorithms, GD and over-relaxed
ADMM; see Fig.~\ref{fig:scheme}. We proved that
precisely the same relation defining lifting of Markov chains 
\eqref{eq:lifting_def}, is satisfied between ADMM and GD. This is
the content of Theorem~\ref{th:lifting}. Although the lifting relation
holds, in general, we cannot guarantee that the matrix $M_A$
associated to ADMM is
a probability transition matrix, since it might have a few negative entries. 
Therefore, in general, Theorem~\ref{th:lifting} is not a Markov chain lifting,
but it is a lifting in a graph theoretical sense.

These negative entries actually 
make this parallel even
more interesting since 
\eqref{eq:conj}, or equivalently \eqref{eq:conjecture_tau},
do not violate theorems of lifted Markov chains
where the square root
improvement is a lower bound, thus the best possible, and
in \eqref{eq:conj} it is an upper bound.
For graphs with low conductance, 
the speedup given by Markov chain lifting is 
negligible. On the other hand, the lifting
between ADMM and GD seems to always give the best possible speedup
achieved by Markov chains, even for graphs with low conductance.
This is numerically confirmed in Fig.~\ref{fig:betas_ring}c. 

Due to the strong analogy with lifted Markov chains and numerical evidence,
we conjectured the upper bound \eqref{eq:conj}, or \eqref{eq:conjecture_tau}, 
which 
was well supported numerically. However,
its formal explanation remains open.

Finally, although we considered a simple class of quadratic objectives, 
when close to the minimum the leading order term of more general convex
functions is usually quadratic. 
In the cases where the dominant term is close to the form
\eqref{eq:general_quad_obj},
the results presented in this paper should still hold.
An attempt to prove our conjecture \eqref{eq:conjecture_tau} 
is under investigation\footnote{Note added: soon after the acceptance of this
paper, we found a proof of \eqref{eq:conjecture_tau} for
a class of quadratic problems. These results will be presented elsewhere.}.

\section*{Acknowledgment}
We  would like to thank Andrea Montanari for his guidance 
and useful discussions. We also would like to thank the anonymous referees
for careful reading of the manuscript and useful suggestions.
This work was partially supported by an NIH/NIAID grant U01AI124302.

\appendix

\section{Proof of Main Results}
\label{sec:proofs}

In the main part of the paper 
we introduced the extended edge set $\hat{\E}$ which essentially duplicates
the edges of the original graph, $|\hat{\E}| = 2 |\E|$.
This is the shortest route to state
our results concisely 
but it complicates the notation in the following proofs.
Therefore, we first introduce the notion of a
factor graph for problem \eqref{eq:general_quad_obj}.

\subsection{Factor Graph}

\begin{figure}
\centering
\includegraphics[scale=1.2]{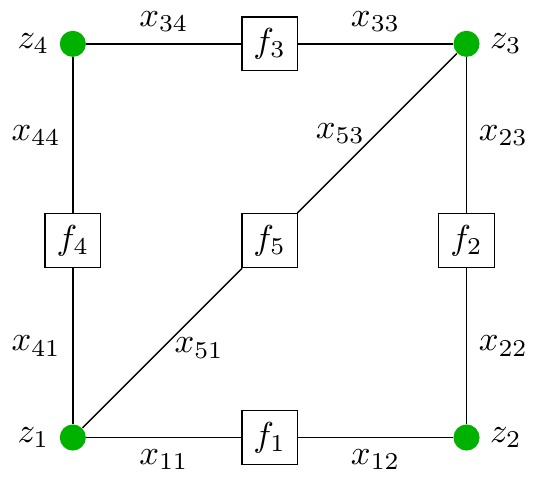}
\caption{Example of a factor graph $\bG$ for problem 
\eqref{eq:general_quad_obj}, and \eqref{eq:fTilde}, where  
$\G$ is the complete graph $K_4$  with one edge removed.
\label{fig:factor_graph}
}
\end{figure}

The factor graph $\bG = (\bF, \bV, \bE)$ for 
problem \eqref{eq:general_quad_obj} is a bipartite graph 
that summarizes how different variables are shared across
different terms in the objective. This is illustrated
in Fig.~\ref{fig:factor_graph}, where for this case 
\eqref{eq:general_quad_obj} is given by
\begin{equation}
\begin{split}
f(\bm{z}) &= q^1(z_1 - z_2)^2 + q^2(z_2 - z_3)^2 + q^3(z_3-z_4)^2 \\
&\qquad +q^4(z_4-z_1)^2 +q^5(z_4-z_2)^2
\end{split}
\end{equation}
while \eqref{eq:fTilde} is given by
\begin{equation}
\begin{split}
&f(\bm{x}) = 
q^1(x_{12} - x_{11})^2  +
q^2(x_{23} - x_{22})^2  \\
&+ q^3(x_{34} - x_{33})^2 + q^4(x_{41} - x_{44})^2 + q^5(x_{45} - x_{52})^2
\end{split}
\end{equation}
where $x_{11} = x_{41}$, $x_{12} = x_{22}$,  
$x_{23} = x_{33}$, and $x_{44}=x_{34}$.

The factor graph $\bG$ has two sets of vertices, $\bF$ and $\bV$.
The circles in Fig.~\ref{fig:factor_graph}
represent the nodes in $\bV=\V$, and the squares
represent the nodes in $\bF = \E$, where $\G=(\V,\E)$ is the original graph. 
Note that each $a \in \bF$ is uniquely associated to one 
edge $e\in \E$, and uniquely associated to one term in the sum of 
the objective.
In equation \eqref{eq:fTilde}
we referred to each term as $f_e = q_e( x_{e_i} - x_{e_j})^2$,
but now we refer to it by $f_a$. 
With a slightly abuse of notation we indiscriminately 
write $a\in \bF$ or $f_a \in \bF$. 
Each node $b \in \bV$ is uniquely associated to one node $i\in \V$,
and uniquely associated to one component of $\vz$. 
Before we referred to this variable by $z_i$, but now we refer to it by $z_b$,
and indiscriminately write $b\in \bF$ or $z_b \in \bF$. 
Each edge $(a,b) \in \bE$ must have $a \in \bF$ and $b \in \bV$,
and its existence implies that the function $f_a$ depends on 
variable $z_b$. Moreover, each edge $(a,b) \in  \bE$ is also 
uniquely associated to one component of $\vx$ in the equivalent 
formulation \eqref{eq:fTilde}. In particular, if $a\in  \bE$ is 
associated to $e \in \E$, and $b \in \bV$ is associated to $i\in \V$, 
then $(a,b) \in \bE$ is
associated to $x_{e_i}$. Here, we denote  $x_{e_i}$ by $x_{ab}$. 
Thus, we can think of $\bE$ as being the same as $\hat{\E}$.
Another way of thinking of $\bE$ and $\vx$ is as follows.
If $(a,b) \in \bE$ then $x_{ab} = z_b$ appears as a constraint
in \eqref{eq:fTilde}. 

Let us introduce the neighbor set of a given node in $\bG$.
For $a \in \bF$, the independent variables of $f_a$ 
are in the set 
\begin{equation}
N_a = \{ b \in \bV: (a,b) \in \bG \}.
\end{equation}
Analogously, for $b \in \bV$, the functions that depend on $z_b$
are in the set 
\begin{equation}
N_b = \{ a \in \bF: (a,b) \in \bG \}.
\end{equation}
In other words,
$N_\bullet$ denotes the neighbors of either circle or square nodes
in $\bG$.
For $a\in \bF$ we define 
\begin{equation}
I_a = \{e \in \bE: e \text{ is incident on } a\}.
\end{equation}
For $b\in \bV$ we define 
\begin{equation}
I_b = \{e \in \bE: e \text{ is incident on } b\}.
\end{equation}

If we re-write problem \eqref{eq:fTilde}  
using this new notation, which indexes variables
by the position they the take on $\bG$, the objective function takes the form
\begin{equation}
f(\bm{x}) = \dfrac{1}{2} \bm{x}^\top Q \bm{x} = 
\dfrac{1}{2} \sum_{a \in \F} \bm{x}_a^\top Q^a \bm{x}_{a} 
\end{equation}
where $Q \in \mathbb{R}^{\bE \times \bE}$ is block diagonal 
and each block, now indexed by
$a \in \bF$, takes the form 
$Q^a = q^a \left( \begin{smallmatrix} +1  & -1 \\ -1 & +1
\end{smallmatrix}\right) $, where $q^a > 0$, and
$\bm{x}_a = (x_{ab}, x_{ac})^\top$
for $(a,b) \in \bE$ and $(a,c) \in \bE$. Here, $q^a$ is the same as 
$q_e$ in the main text. We also have the constraints
$x_{a b} = x_{a'b} = z_b $ for each $a, a' \in N_b$ and $b \in \bV$.
The row stochastic matrix $S$ introduced in the ADMM iterations
is now expressed as $S\in \mathbb{R}^{|\bE|\times|\bV|}$
and has a single $1$ per row such that $S_{eb} = 1$ if and only if edge
$e\in \bE$ is incident on $b\in \bV$.
Notice that $S^\top S = \Dd$ is the degree matrix of the original graph $\G$.

With this notation at hands, we now proceed to the proofs.

\subsection{Proof of Theorem~\ref{th:linearADMM}}
Recall that $B = S (S^\top D_{\rho} S)^{-1} S^\top D_{\rho}$, thus
$B^2 = B$ is a projection operator, and 
$B^\perp = I - B$ its orthogonal complement. 
Consider updates \eqref{eq:admm_matrix}.
Substituting $\vx^{t+1}$ and $\vm^{t+1}$ into the other variables
we obtain
\begin{equation*}
\begin{pmatrix}
I & 0 & 0 \\
I & I & 0 \\
-I & I & I \\
\end{pmatrix}
\begin{pmatrix}
\vs^{t+1} \\
\vu^{t+1} \\
\vn^{t+1}
\end{pmatrix}
=
\begin{pmatrix}
(1-\gamma) I & B & \gamma BA \\
(1-\gamma) I & I & \gamma A \\
0 & 0 & 0
\end{pmatrix}
\begin{pmatrix}
\vs^{t} \\
\vu^{t} \\
\vn^{t}
\end{pmatrix}
\end{equation*}
which can be easily inverted yielding
\begin{align}
\vs^{t+1} &= (1-\gamma)\vs^t + B\vu^t + \gamma BA\vn^t \label{eq:uup1}, \\
\vu^{t+1} &= B^\perp \vu^t + \gamma B^\perp A\vn^t \label{eq:uup2}, \\
\vn^{t+1} &= (1-\gamma)\vs^t + (B - B^\perp)\vu^t + 
\gamma (B - B^\perp)A\vn^t. \label{eq:uup3}
\end{align}
Note the following important relations:
\begin{align}
B \vn^t & = \vs^t,  & B^\perp \vn^t &= - \vu^t, \label{eq:rel1} \\
B \vs^t &= \vs^t,  & B^\perp \vs^t &= 0, \label{eq:rel2}\\
B^\perp \vu^t &= \vu^t,  & B \vu^t &= 0. \label{eq:rel3}
\end{align}
Equation \eqref{eq:rel2} is a simple consequence of the definition of $B$, i.e.
\begin{equation}
B \, \vs^t = S (S^\top D_{\vrho} S)^{-1} ( S^\top D_{\vrho} S)\vz^t = \vs^t,
\end{equation}
which also implies $B^\perp \vs^t = \vzero$.
Since
$B B^\perp = 0$, acting with $B$ over \eqref{eq:uup2} implies 
$B \vu^{t} = \vzero$ for
every $t$, and also $B^\perp \vu^t = \vu^t$, which shows \eqref{eq:rel3}. 
Now \eqref{eq:rel1} follows from these facts and the 
own definition
$\vn^t = \vs^t - \vu^t$.
Finally, applying \eqref{eq:rel1} on 
\eqref{eq:uup3} we obtain $\vn^{t+1} = T_A \vn^{t}$ where
$T_A = I - \gamma (A + B - 2BA)$.

\subsection{Proof of Theorem~\ref{th:TG_stochastic}}
Write $Q = Q^{+} + Q^{-}$ where $Q^+$ is diagonal and has only positive
entries, and $Q^-$ only has off-diagonal and negative entries.
First, notice that $(S^\top Q^+ S)$ is also diagonal. Indeed, for
$b \in \bV$ and $c \in \bV$,
$(S^\top Q^+ S)_{b c} = 
\sum_{e \in \bE} S_{e b} Q^+_{ee} S_{e c} = 
\delta_{bc} \sum_{e \in I_b} Q^+_{ee} $
where $\delta$ is the Kronecker delta.
By a similar argument,
$S^\top Q^- S$ is off-diagonal.
Hence, if $b \neq c$ we have
\begin{equation}\label{eq:tg_pos}
(T_G)_{bc} = 
-\alpha \sum_{ e \in I_b} \sum_{e' \in I_c} Q^-_{ee'} \ge 0 .
\end{equation}
Recall that $M_G = (I - D_G)^{-1} (T_G - D_G)$, where $D_G \ne I$ is
diagonal. 
For $M_G$ to be non-negative we first impose that $(D_G)_{bb} < 1$ 
for all $b\in \bV$. 
Then, since the off-diagonal elements of $T_G$ are automatically positive by
\eqref{eq:tg_pos}, we just
need to consider the diagonal elements of $T_G - D_G$. Thus we
require that for every $b \in \bV$,
\begin{equation}
1 - \alpha \sum_{e \in I_b} Q_{ee} + (D_G)_{bb} \ge 0.
\end{equation}
Denoting 
$Q_{\textnormal{max}} = \max_{b \in \bV} \sum_{e \in I_b} Q_{ee} $ and
$D_{G,\textnormal{min}}$ the smallest element of $D_G$,
the matrix 
$M_G$ will be non-negative 
provided $\alpha \le (1 + D_{G,\textnormal{min}})/Q_{\textnormal{max}}$.

Notice that $S\bm{1}_{|\bV|} = \bm{1}_{|\bE|}$ 
and $Q \bm{1} = \bm{0}$. Thus
$S^\top Q S \bm{1} = \bm{0}$, implying $T_G \bm{1} =
\bm{1}$, and 
$\bm{1}^\top T_G = \bm{1}^\top$. From this we have
$M_G \bm{1} = \bm{1}$ and $\bm{1}^\top M_G = \bm{1}^\top$, so all the rows
and columns of $M_G$ sum up to one.

\subsection{Proof of Lemma~\ref{th:TAoneisone}}
We proved above that $M_G$ is a doubly stochastic matrix. Now let
us consider $M_A$. Recall the definition of
$B = S^\top (S^\top D_{\vrho} S)^{-1} S^\top D_{\vrho}$.
Note that the action of $B$ on a vector $\vv \in \mathbb{R}^{|\bE|}$ 
is to take a weighted
average of its components, namely, if $(a,b) \in \bE$ then
\begin{equation}
(\B \bm{v})_{ab} = 
\dfrac{\sum_{c \in N_b} \rho_{cb}v_{cb}}{ \sum_{c\in N_b} \rho_{cb} }.
\end{equation}
Therefore, 
$\B {\bm 1} = {\bm 1}$. Recall that $Q \bm{1} = \bm{0}$, thus
$\A {\bm 1} = \bm{1}$, where $A = (I + D_{\vrho}^{-1}Q)^{-1}$, which
implies $T_A\bm{1} = \bm{1}$, and in turn $M_A\bm{1} = \bm{1}$. 
Now the other relations follow trivially.

\subsection{Proof of Theorem~\ref{th:lifting}}
Due to the block diagonal structure of $Q$ it is possible write $A$  
explicitly as
\begin{equation}\label{eq:F_rho}
A = I - F Q,
\end{equation}
where $F$ is a block diagonal matrix with $|\bF|$ blocks.
Each block $F^a$, for $a\in \bF$, is of the form
\begin{equation}
\qquad
F^a = \dfrac{q^{a}}{\rho_{ab}\rho_{ac} + q^{a}(\rho_{ab} + \rho_{ac})}
\begin{pmatrix}
\rho_{ac} & 0 \\
0 & \rho_{ab}
\end{pmatrix},
\end{equation}
where $b,c \in N_a$.
Now by the definition of $B$ we have
$S^\top D_{\bm{\rho}} B = S^\top D_{\bm{\rho}}$. Hence,
\begin{align}
S^\top D_{\bm{v}_A} M_A S &= S^\top D_{\bm{\rho}}(I - D_A) S
- \gamma S^\top D_{\bm{\rho}} F Q S, \label{eq:lift1}\\
D_{\bm{v}_G} M_G &= (I - D_G) - \alpha S^\top Q S. \label{eq:lift2}
\end{align}
Equating
the first term of \eqref{eq:lift1} to the first term of
\eqref{eq:lift2}, and also the second terms to each other, on 
using \eqref{eq:F_rho} we obtain
\begin{align}
&S^\top D_{\vrho} (I-D_A) S = I - D_G, \\
&\alpha = \dfrac{
\gamma \, q^a \rho_{ab} \rho_{ac}
}{
\rho_{ab} \rho_{ac} + q^a( \rho_{ab} + \rho_{bc} )
}, \label{eq:alpha_rho_proof}
\end{align}
where \eqref{eq:alpha_rho_proof} must hold for all $a \in \bF$
and $b,c \in N_a$. This gives the second equality in
\eqref{eq:liftingMAMG} together with relations \eqref{eq:lifting_params}
and \eqref{eq:rho_alpha}.
Finally, since diagonal matrices commute,
$S^\top\bm{v}_A = S^\top (I-D_A) D_{\bm{\rho}} S\bm{1}_{|\bV|} = 
(I-D_G) \bm{1}_{|\bV|} = \bm{v}_G$, which gives the first relation
in \eqref{eq:liftingMAMG}.

\subsection{Proof of Theorem~\ref{th:negative}}
It suffices to show one example with at least one negative entry. 
Let $\G$ be the complete graph $K_4$ with one edge
removed, as shown in Fig.~\ref{fig:factor_graph}.
By direct inspection one finds the following sub-matrix of
$T_A$:
\begin{equation}\label{eq:TA_sub}
T^{(S)} = 
\begin{pmatrix}
(T_A)_{21} &  (T_A)_{24} \\
(T_A)_{31} &  (T_A)_{34} 
\end{pmatrix}
\end{equation}
whose elements are explicitly given by
\begin{align}
(T_A)_{21} &= 
\tfrac{\gamma \, \rho_{11}(\rho_{12}- \rho_{22})}
{(\rho_{12} + \rho_{22})(\rho_{11} + \rho_{12} + \rho_{11}\rho_{12})} ,
\label{eq:TA21} \\
(T_A)_{24} &= 
\tfrac{2 \gamma }
{(\rho_{12} + \rho_{22})\big(1+\rho_{22}^{-1} + \rho_{23}^{-1}\big)} ,
\label{eq:TA24} \\
(T_A)_{31} &= 
\tfrac{2 \gamma}
{(\rho_{12} + \rho_{22})\big(1+\rho_{11}^{-1} + \rho_{12}^{-1}\big)} ,
\label{eq:TA31} \\
(T_A)_{34} &= 
- \tfrac{ \gamma \, \rho_{23}(\rho_{12}- \rho_{22})}
{(\rho_{12} + \rho_{22})(\rho_{22} + \rho_{23} + \rho_{22}\rho_{23})} .
\label{eq:TA34} 
\end{align}
First notice that subtracting $D_A$ from $T_A$ does not affect $T^{(S)}$.
Now recall that all components of $\vrho$ must be strictly positive. 
The elements \eqref{eq:TA21} and \eqref{eq:TA34} have opposite signs, so one
of them is negative. Since \eqref{eq:TA24} and \eqref{eq:TA31}
are both positive,
one cannot remove the negative entries of an entire row of $T_A$ by 
multiplying $T_A$ by the diagonal matrix $(I - D_A)^{-1}$. 
Therefore, $M_A = (I - D_A)^{-1}(T_A - D_A)$ has
at least one negative entry.


\end{document}